\documentclass{article}

\usepackage{microtype}
\usepackage{graphicx}
\usepackage{subfigure}
\usepackage{booktabs} 

\usepackage{hyperref}

\usepackage[accepted]{icml2025}

\usepackage{amsmath}
\usepackage{amssymb}
\usepackage{mathtools}
\usepackage{amsthm}

\usepackage[capitalize,noabbrev]{cleveref}

\usepackage{amsmath} 
\usepackage{multirow}
\usepackage{tikz}
\usepackage{todonotes}
\usepackage{booktabs}
\usetikzlibrary{arrows.meta}
\usepackage{amsmath}

\newcommand{\size}[2]{{\fontsize{#1}{0}\selectfont#2}}

\theoremstyle{plain}

\theoremstyle{definition}

\theoremstyle{remark}

\DeclareMathOperator{\Tr}{Tr}

\icmltitlerunning{}

\begin{document}

\twocolumn[
\icmltitle{Learning from Linear Algebra: A Graph Neural Network Approach to \\ Preconditioner Design for Conjugate Gradient Solvers}




\begin{icmlauthorlist}
\icmlauthor{Vladislav Trifonov}{skoltech,sber}
\icmlauthor{Alexander Rudikov}{airi,skoltech}
\icmlauthor{Oleg Iliev}{itwm}\\
\icmlauthor{Yuri M. Laevsky}{ras}
\icmlauthor{Ivan Oseledets}{airi,skoltech}
\icmlauthor{Ekaterina Muravleva}{sber,skoltech}
\end{icmlauthorlist}

\icmlaffiliation{skoltech}{Skolkovo Institute of Science and Technology, Moscow, Russia}
\icmlaffiliation{sber}{Sberbank of Russia, AI4S Center, Moscow, Russian Federation}
\icmlaffiliation{airi}{Artificial Intelligence Research Institute (AIRI), Moscow, Russia}
\icmlaffiliation{itwm}{Fraunhofer Institute for Industrial Mathematics ITWM, Kaiserslautern, Germany}
\icmlaffiliation{ras}{Institute of Computational Mathematics and Mathematical Geophysics SB RAS, Novosibirsk, Russia}

\icmlcorrespondingauthor{Vladislav Trifonov}{vladislav.trifonov@skoltech.ru}

\icmlkeywords{Machine Learning, ICML}

\vskip 0.3in
]



\printAffiliationsAndNotice{}  

\begin{abstract}
Large linear systems are ubiquitous in modern computational science and engineering. The main recipe for solving them is the use of Krylov subspace iterative methods with well-designed preconditioners. Recently, GNNs have been shown to be a promising tool for designing preconditioners to reduce the overall computational cost of iterative methods by constructing them more efficiently than with classical linear algebra techniques. Preconditioners designed with these approaches cannot outperform those designed with classical methods in terms of the number of iterations in CG. In our work, we recall well-established preconditioners from linear algebra and use them as a starting point for training the GNN to obtain preconditioners that reduce the condition number of the system more significantly than classical preconditioners. Numerical experiments show that our approach outperforms both classical and neural network-based methods for an important class of parametric partial differential equations. We also provide a heuristic justification for the loss function used and show that preconditioners obtained by learning with this loss function reduce the condition number in a more desirable way for CG.
\end{abstract}

\section{Introduction}

Modern computational science and engineering problems are often based on parametric partial differential equations (PDEs). The lack of analytical solutions for realistic engineering problems (heat transfer, fluid flow, structural mechanics, etc.) leads researchers to exploit advances in numerical analysis. The basic numerical methods for solving PDEs, such as finite element, finite difference, finite volume and meshless methods (e.g., smoothed particle hydrodynamics), result in a system of linear equations $Ax = b, \,\, A \in \mathbb{R}^{n \times n}$, ~$x \in \mathbb{R}^n$, and $b \in \mathbb{R}^n$. These systems are usually sparse, i.e. the number of non-zero elements is $\ll n^2$. Furthermore, some classes of parametric PDEs are characterized by a very large dimension of the parametric space and by a high variation of the parameters for a given sample.

Typically, the application of parametric PDEs produces large linear systems, often with entries of varying scale, and therefore poses significant computational challenges. Projection Krylov subspace iterative methods are widely used to solve such systems. They rely on finding an optimal solution in a subspace constructed as follows: $\mathcal{K}_r(A, b) = \text{span}\{ b, ~Ab, ~A^2b, \dots, ~A^{r-1}\}$.

The conjugate gradient (CG) method is used to solve large sparse systems with symmetric and positive definite (SPD) matrices~\citep{saad2003iterative, axelsson1996iterative}. CG has a well-established convergence theory and convergence guarantees for any symmetric matrix. However, the convergence rate of CG is determined by $\sqrt{\kappa(A)}$, which is typically large. The condition number $\kappa(A)$ of an SPD matrix $A$, defined in $2\text{-norm}$, is a ratio between the maximum and minimum eigenvalues $\kappa (A) = |\lambda_{\max}| \big/ |\lambda_{\min}|$.

Real-world applications with non-smooth high-contrast coefficient functions and high-dimensional linear systems separate eigenvalues and results into ill-conditioned problems with high $\kappa(A)$. Decades of research in numerical linear algebra have been devoted to constructing preconditioners $P$ for ill-conditioned $A$ to improve the condition number in the form (for left-preconditioned systems) $\kappa(P^{-1}A) \ll \kappa(A)$.

The well-designed preconditioner should tend to approximate $A$, be easily invertible and be sparse or admit an efficient matrix-vector product. The construction of a preconditioner is typically a trade-off between the quality of the approximation and the cost of storage/inversion of the preconditioner \citep{saad2003iterative}.

In this paper, we propose a novel neural method of preconditioner design for symmetric positive definite matrices called PreCorrector (short for Preconditioner Corrector). Preconditioners constructed with PreCorrector have better effect on the spectrum than classical preconditioners by learning correction for the latter. Our contributions are as follows:

\begin{itemize}
    \item[1.] We propose a novel scheme for preconditioner design based on learning correction for well-established preconditioners from linear algebra with the GNN, which has better effect on spectrum than classical preconditioners.
    \item[2.] We propose an understanding of the loss function used with emphasis on low frequencies. We also provide experimental justification for the understanding of learning with such an objective.
    \item[3.] We propose a novel dataset generation approach with a measurable complexity metric that addresses real-world problems.
    \item[4.] We provide extensive experiments with varying matrix sizes and dataset complexities to demonstrate the superiority of the proposed approach and loss function over classical preconditioners.
\end{itemize}

\section{Neural design of preconditioner}
\label{sec:neural_prec_design}

\paragraph{Related work}
While there are a dozen different preconditioners in linear algebra, for example \citep{saad2003iterative, axelsson1996iterative}: block Jacobi preconditioner, Gauss-Seidel preconditioner, sparse approximate inverse preconditioner, algebraic multigrid methods, etc., the choice of preconditioner depends on the specific problem, and practitioners often rely on a combination of theoretical understanding and numerical experimentation to select the most effective preconditioner. Even a brief description of all of them is beyond the scope of a single research paper. One can refer to the related literature for more details

The growing popularity of neural operators for learning mappings between infinite dimensional spaces (e.g., \citep{hao2024newton, cao2024laplace, raonic2024convolutional}) is also present in recent work on using neural networks to speed up iterative solvers. The FCG-NO~\citep{rudikov2024neural} approach combines neural operators with the conjugate gradient method to act as a nonlinear preconditioner for the flexible conjugate gradient method~\citep{notay2000flexible}. This method uses a proven convergence bound as a training loss. A novel class of hybrid preconditioners~\citep{kopanivcakova2024deeponet} combines DeepONet with standard iterative methods to solve parametric linear systems. This framework uses DeepONet for low-frequency error components and conventional methods for high-frequency components. The HINTS~\citep{zhang2022hybrid} method integrates traditional relaxation techniques with DeepONet. It targets different spectral regions, ensuring a uniform convergence rate. It is also possible to use convolutional neural networks to speed up multigrid method~\citep{azulay2022multigrid, li2024machine}, which require materialization of sparse matrices into dense format. However, these approaches can suffer from the curse of dimensionality when applied to large linear systems and can be too expensive to apply at each iteration step.

The authors of \citep{li2023learning, hausner2023neural} present a novel approach to preconditioner design using GNNs that aim to approximate the matrix factorization and use it as a preconditioner. These approaches use shallow GNNs and typically require a single inference before the iteration process. GNNs take the initial left hand side matrix and right hand side vector as input and construct preconditioners in the form of a Cholesky decomposition. However, these GNNs cannot produce preconditioners that have a better effect on the condition number of the solving system than their classical analogues.

\paragraph{Problem statement}
We consider systems of linear algebraic equations from the discretization of differential operators $Ax = b$ formed with a symmetric positive definite (SPD) matrix $A \succ 0$. One can use Gaussian elimination of complexity $\mathcal{O}(n^3)$ to solve small linear systems, but not real-world problems, which produce large and ill-conditioned systems.


\paragraph{Preconditioned linear systems}
Before solving initial systems by iterative methods, we want to obtain a preconditioner $P$ such that the preconditioned linear system $P^{-1}Ax = P^{-1}b$ has a lower condition number than the initial system. If one knows the sparsity pattern of $A$, then possible options are incomplete LU decompositions (ILU)~\citep{saad2003iterative}: (i) with $p$-level of fill-in denoted as ILU($p$) and (ii) ILU decomposition with threshold with $p$-level of fill-in denoted as ILUt($p$). Additional information about these preconditioners can be found in the Appendix~\ref{app:ilu}.

In this paper we focus on the SPD matrices so instead of ILU, ILU($p$) and ILUt($p$) we use the incomplete Cholesky factorization IC, IC($p$) and ICt($p$). Further, we will form the preconditioners in the form of Cholesky decomposition~\citep{trefethen2022numerical} $P = LL^\top$ with sparse $L$ obtained by different methods. 

\paragraph{Preconditioners with neural networks} 
Our ultimate goal is to find such a decomposition that $\kappa((L(\theta)L(\theta)^\top)^{-1}A) \ll \kappa((LL^\top)^{-1}A) \ll \kappa(A)$, where $L$ is the classical numerical IC decomposition and $L(\theta) = \mathcal{F}(A)$ is an approximate decomposition with some function $\mathcal{F}$. Several papers~\citep{li2023learning, hausner2023neural} suggest using GNN as a function $\mathcal{F}$ to minimize certain loss function:
\begin{equation}
    \label{eq:F_as_GNN}
    L(\theta) = \text{GNN}(\theta, A, b)\, .
\end{equation}
\paragraph{Loss function}

The key question is which objective function to minimize in order to construct a preconditioner. A natural choice, which is also used in \citep{hausner2023neural}, is: 
\begin{equation}
    \label{eq:loss_minus_A}
     \min \big \Vert P - A \big \Vert_F^2.
\end{equation}

By design, this objective minimizes high frequency components (large eigenvalues), which is not desired. Low frequency components (small eigenvalues) are the most important because they correspond to the simulated phenomenon, when high frequency comes from discretization methods. It is also known that CG eliminates errors corresponding to high frequencies first and struggles the most with low frequencies. We suggest using $A^{-1}$ as the weight for the previous optimization objective to take into account low frequency since $\lambda(A) = \lambda^{-1}(A^{-1})$:
\begin{equation}
    \label{eq:loss_minus_A_invA}
    \min \big \Vert (P - A) A^{-1} \big \Vert^2_F~.
\end{equation}

Let us rewrite this objective using Hutchinson's estimator \citep{hutchinson1989stochastic}:
\begin{multline}
    \label{eq:deriv_loss}
    \big \Vert \big(P - A\big) A^{-1} \big \Vert^2_F = \big \Vert P A^{-1} - I \big \Vert_F^2 = \\
    \Tr \Big( (PA^{-1} - I)^\top (PA^{-1} - I) \Big) = \\
    \mathbb{E}_\varepsilon \Big[ \varepsilon^\top(PA^{-1} - I)^\top (PA^{-1} - I)\varepsilon \Big] = \\ 
    \mathbb{E}_\varepsilon \big\Vert (PA^{-1} - I)\varepsilon \big\Vert_2^2, \quad \varepsilon \sim \mathcal{N}(0,1).
\end{multline}

Suppose we have a dataset of linear systems $A_i x_i = b_i$, then the training objective with $\varepsilon = b_i, ~P = L(\theta)L(\theta)^\top$ and $A^{-1}_ib_i = x_i$ will be:
\begin{equation}
    \label{eq:loss_with_rhs}
    \mathcal{L} = \frac{1}{N} \sum_{i=1}^{N} \big \Vert L(\theta)L(\theta)^\top x_i - b_i \big\Vert_2^2~.
\end{equation}

This loss function has appeared previously in related research~\citep{li2023learning} but with an understanding of the inductive bias from the PDE data distribution. We claim that training with loss~(\ref{eq:loss_minus_A_invA}) allows to obtain better preconditioners than with loss~(\ref{eq:loss_minus_A}). In the Section~\ref{sec:Experiments}, we demonstrate that loss~(\ref{eq:loss_minus_A_invA}) does indeed mitigate low-frequency components.

\section{Learn correction for ILU}
\label{sec:learn_corrrection_ilu}

Our main goal is to construct preconditioners that reduce the condition number of an SPD matrix more than classical preconditioners with the same sparsity pattern. Several previous papers suggested to use convolutional neural networks with sparse convolutions~\cite{sappl2019deep, cali2023neural} and graph neural networks~\cite{li2023learning, hausner2023neural}. However, none of the approaches, to the best of authors knowledge, did not achieve to obtain better effect of spectrum with neural-preconditioner construction while maintaining the same sparsity level. We propose to learn corrections for classical preconditioner to achieve such effect.
\begin{figure}[ht]
\vskip 0.2in
\begin{center}
\centerline{\includegraphics[width=\columnwidth]{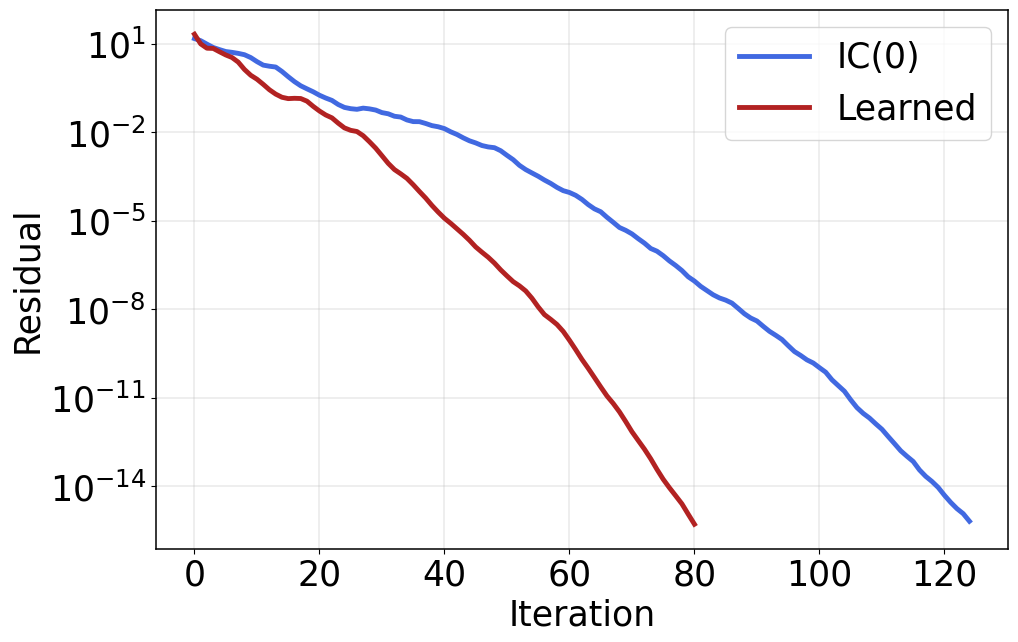}}
\caption{Inplace updating of IC($0$) factor allows to obtain better preconditioner.}
\label{fig:inplace_upd}
\end{center}
\vskip -0.2in
\end{figure}

\paragraph{Inplace ILU update}
PreCorrector is built around the idea that classical preconditioners of the ILU family produce reliable preconditioners for the CG method. Although the algorithm for their construction is known, they do not exist uniquely and one can try to train neural network to construct preconditioners of the same sparsity pattern with better effect on spectrum. This can be shown by inplace updating of IC(0) factors with gradient descent (Figure~\ref{fig:inplace_upd}).

In this approach we treat each element of the sparse matrix as a neural network parameter and update an already efficient preconditioner. We follow the idea of recent research and use GNN to parametrize preconditioner learning. 

\paragraph{PreCorrector}
The duality between sparse matrices and graphs is used to obtain vertices and edges, such as $Ax=b \rightarrow \mathcal{G} = (\mathcal{V}, \mathcal{E})$, where $a_{i,j} = e_{i,j} \in \mathcal{E}, b_i = v_i \in \mathcal{V}$. 

Instead of passing left hand side matrix $A$ as input to GNN in~\eqref{eq:F_as_GNN}, we propose: (i) to pass $L$ from the IC decomposition to the GNN and (ii) to train GNN to predict a correction for this decomposition:
\begin{equation}
    \label{eq:correct_ic0}
    L(\theta) = L + \alpha \cdot \text{GNN}(\theta, L)~.
\end{equation}

The correction coefficient $\alpha$ is also a learning parameter that is updated during training. At the beginning of training, we set $\alpha = 0$ to ensure that the first gradient updates come from pure IC factorization. Moreover, GNN in~(\ref{eq:correct_ic0}) takes as input the lower-triangular matrix $L$ from IC instead of $A$, so we are not anchored to a single specific sparse pattern of $A$ and we can: (i) omit half of the graph and speed up the training process and (ii) use different sparsity patterns to obtain even better preconditioners. In Experiment section, we show that the proposed approach with input $L$ from IC(0) and ICt(1) produces better preconditioners compared to classical IC($0$) and ICt($1$) and previous neural-designed preconditioners.

\paragraph{Graph neural network architecture}
We use ubiquitous encode-process-decode~\cite{battaglia2018relational} architecture for the GNN, where multilayer perceptrons are used for encoders and decoders. Process-block consists of message-passing layer that increase receptive field of the GNN with multiple rounds. Before encoder and after decoder we apply normalization and renormalization for the edges respectively. One can see PreCorrector's forward pass in Algorithm~\ref{alg:precorrector_forward}.

Note that in~\eqref{eq:correct_ic0} we omit the right-hand side vector $b$ and do not use it as input for GNN. In our experiments, we observe that GNN with the right-hand side as node input does not effect resulting preconditioner quality, while it suffers from sensitivity to the right-hand side vector. Although we can either use a different representation of the node features (e.g., as in~\cite{hausner2023neural}) or use diagonal elements of the matrix. Instead we set vector $\textbf{1} = [1, \dots, 1]^T$ as the node input to the message passing layer, which allowed us to reduce the number of parameters by a factor of $2.3$ while maintaining the quality of the resulting preconditioner. Details on the architecture of the PreCorrector architecture can be found in the Appendix~\ref{app:architecture}.

\section{Dataset}
\label{sec:datasets}

We test PreCorrector on SPD matrices obtained by discretization of elliptic equations. We consider a 2D diffusion equation:
\begin{equation}
    \label{eq:diffusion_equation}
    \begin{split}
    -&\nabla \cdot \big(k(x)\nabla u(x)\big) = f(x), ~\text{in}~\Omega\\
     &u(x)\Big|_{x\in \partial{\Omega}} = 0
    \end{split} \,\,\, ,
\end{equation}
and 2D Poisson equation:
\begin{equation}
    \label{eq:poisson_equation}
    \begin{split}
    -&\nabla^2 u(x) = f(x), ~\text{in}~\Omega\\
     &u(x)\Big|_{x\in \partial{\Omega}} = 0
    \end{split} \,\,\, ,
\end{equation}
where $k(x)$ is a diffusion coefficient, $u(x)$ is a solution and $f(x)$ is a forcing term. 

The diffusion equation is chosen because it occurs frequently in many engineering applications, such as: composite modeling~\citep{carr2016semi}, geophysical surveys~\citep{oristaglio1984diffusion}, fluid flow modeling~\citep{muravleva2021multigrid}. In these industrial applications, the coefficient functions are discontinuous, i.e. they change rapidly within neighbouring cells. Examples also include the flow of immiscible fluids of different viscosities and fluid flow in heterogeneous porous media.

The condition number of a linear system depends on both the grid size and the contrast, but usually in scientific machine learning research, high contrast is not taken into account. In Section~\ref{sec:comparison} we demonstrate that the previous approach can handle growing matrix size with constant coefficients quite well, but faces problems with growing contrast in the coefficients. 

We propose to use a Gaussian Random Field (GRF) $\phi(x)$ to generate the coefficients. In our experiments we use the \texttt{parafields} library\footnote{https://github.com/parafields/parafields} to generate GRFs. To control the complexity of the diffusion equation with discontinuous coefficients, we can measure the contrast in the GRF:
\begin{equation}
    \label{eq:contrast}
    \text{contrast} = \exp{\big( \max{\big(\phi(x)\big)} - \min{\big(\phi(x)\big)} \big)}.
\end{equation}
Then we generate coefficients for the equation~(\ref{eq:diffusion_equation}) as $k(x) = \exp{(\phi(x))}$. By changing the variance in the GRF, we can control the contrast of the coefficients and thus the complexity. More details about datasets can be found in the Appendix~\ref{app:dataset}.

\begin{table*}[!h]
    \footnotesize
    \caption{Comparison on diffusion equation with variance $1.1$: classical algorithms IC($0$) and ICt($1$) and PreCorrector. Lower is better, the best results are bold. Pre-time stands for precomputations time.}
    \label{table:diff_1.1}
    \centering
    \begin{tabular}{@{}lcccccc@{}} 
        \\\toprule
        &  &  & Time (iters) & Time (iters) & Time (iters) & Time (iters) \\
        Grid  & Method  & Pre-time &  to $10^{-3}$  &to $10^{-6}$ &to $10^{-9}$  &to $10^{-12}$  \\
        \midrule
        \multirow{9.5}{*}{$128\times128$}&  \multirow{2}{*}{$\text{IC}(0)$}  & \multirow{2}{*}{$4.6\cdot10^{-4}\pm9.4\cdot10^{-6}$} & $1.267\pm0.154$ & $1.573\pm0.186$ & $1.816\pm0.218$ & $2.018\pm0.243$ \\ 
        &  &  & \size{8}{($171\pm6.8$)} & \size{8}{($213\pm7.3$)} & \size{8}{($245\pm7.0$)} & \size{8}{($273\pm6.8$)} \\\cmidrule(r){3-7}
        &  \multirow{2}{*}{PreCor$\big[\text{IC}(0)\big]$}  & \multirow{2}{*}{$8.4\cdot10^{-4}\pm1.3\cdot10^{-4}$} & $\textbf{0.763}\pm0.142$ & $\textbf{0.969}\pm0.180$ & $\textbf{1.164}\pm0.218$ & $\textbf{1.355}\pm0.254$ \\ 
        &  &  & \size{8}{($108\pm10.3$)} & \size{8}{($138\pm13.4$)} & \size{8}{($166\pm16.5$)} & \size{8}{($193\pm19.6$)} \\\cmidrule(r){2-7}
        &  \multirow{2}{*}{$\text{ICt}(1)$}  & \multirow{2}{*}{$3.1\cdot10^{-3}\pm6.2\cdot10^{-5}$} & $0.839\pm0.121$ & $1.040\pm0.148$ & $1.201\pm0.171$ & $1.337\pm0.190$ \\ 
        &  &  & \size{8}{($104\pm4.1$)} & \size{8}{($129\pm4.5$)} & \size{8}{($150.0\pm4.2$)} & \size{8}{($166\pm4.0$)} \\\cmidrule(r){3-7}
        &  \multirow{2}{*}{PreCor$\big[\text{ICt}(1)\big]$}  & \multirow{2}{*}{$3.6\cdot10^{-3}\pm2.3\cdot10^{-4}$} & $\textbf{0.450}\pm0.039$ & $\textbf{0.572}\pm0.042$ & $\textbf{0.687}\pm0.049$ & $\textbf{0.798}\pm0.054$ \\ 
        &  &  & \size{8}{($63\pm3.8$)} & \size{8}{($80\pm4.5$)} & \size{8}{($96\pm5.2$)} & \size{8}{($112\pm5.8$)} \\
        \midrule
        \multirow{9.5}{*}{$256\times256$}&  \multirow{2}{*}{$\text{IC}(0)$}  & \multirow{2}{*}{$1.8\cdot10^{-3}\pm6.0\cdot10^{-5}$} & $8.638\pm2.000$ & $10.635\pm2.355$ & $12.249\pm2.654$ & $13.608\pm2.905$ \\
        &  &  & \size{8}{($345\pm13.5$)} & \size{8}{($425\pm14.9$)} & \size{8}{($490\pm14.9$)} & \size{8}{($544\pm14.4$)} \\\cmidrule(r){3-7}
        &  \multirow{2}{*}{PreCor$\big[\text{IC}(0)\big]$}  & \multirow{2}{*}{$3.3\cdot10^{-3}\pm6.1\cdot10^{-4}$} & $\textbf{3.388}\pm0.544$ & $\textbf{4.199}\pm0.663$ & $\textbf{4.943}\pm0.775$ & $\textbf{5.637}\pm0.875$ \\ 
        &  &  & \size{8}{($144\pm12.6$)} & \size{8}{($179\pm15.5$)} & \size{8}{($211\pm17.7$)} & \size{8}{($241\pm19.6$)} \\\cmidrule(r){2-7}
        &  \multirow{2}{*}{$\text{ICt}(1)$}  & \multirow{2}{*}{$1.2\cdot10^{-2}\pm1.8\cdot10^{-4}$} & $5.278\pm0.593$ & $6.515\pm0.709$ & $7.511\pm0.809$ & $8.340\pm0.890$ \\ 
        &  &  & \size{8}{($209\pm8.4$)} & \size{8}{($259\pm9.0$)} & \size{8}{($298\pm9.1$)} & \size{8}{($332\pm8.8$)} \\\cmidrule(r){3-7}
        &  \multirow{2}{*}{PreCor$\big[\text{ICt}(1)\big]$}  & \multirow{2}{*}{$1.4\cdot10^{-2}\pm6.8\cdot10^{-4}$} & $\textbf{2.719}\pm0.381$ & $\textbf{3.378}\pm0.462$ & $\textbf{3.976}\pm0.539$ & $\textbf{4.534}\pm0.608$ \\ 
        &  &  & \size{8}{($105\pm8.3$)} & \size{8}{($131\pm9.8$)} & \size{8}{($154\pm11.2$)} & \size{8}{($176\pm12.4$)} \\
        \bottomrule
    \end{tabular}
\end{table*}

\begin{table*}[!h]
    \small
    \caption{Comparison on diffusion equation with variance $1.1$: classical algorithm ICt($5$) and PreCor$\big[\text{ICt}(1)\big]$. Lower is better, the best results are bold. Pre-time stands for precomputations time.}
    \label{table:diff_1.1_ict5}
    \centering
    \begin{tabular}{@{}lcccccc@{}} 
        \\\toprule
        &  &  & Time (iters) & Time (iters) & Time (iters) & Time (iters) \\
        Grid  & Method  & Pre-time &  to $10^{-3}$  &to $10^{-6}$ &to $10^{-9}$  &to $10^{-12}$ \\
        \midrule
        \multirow{4.5}{*}{$128\times128$}& \multirow{2}{*}{$\text{ICt}(5)$}  & \multirow{2}{*}{$7.5\cdot10^{-3}\pm3.4\cdot10^{-4}$} & $0.716\pm0.054$ & $0.890\pm0.062$ & $1.033\pm0.070$ & $1.152\pm0.075$ \\
        &  &  & \size{8}{($52\pm2.1$)} & \size{8}{($65\pm2.1$)} & \size{8}{($76\pm2.0$)} & \size{8}{($84\pm2.0$)} \\\cmidrule(r){2-7}
        &  \multirow{2}{*}{PreCor$\big[\text{ICt}(1)\big]$}  & \multirow{2}{*}{$3.6\cdot10^{-3}\pm2.3\cdot10^{-4}$} & $\textbf{0.450}\pm0.034$ & $\textbf{0.572}\pm0.042$ & $\textbf{0.687}\pm0.049$ & $\textbf{0.798}\pm0.054$ \\ 
        &  &  & \size{8}{($63\pm3.8$)} & \size{8}{($80\pm4.5$)} & \size{8}{($96\pm5.2$)} & \size{8}{($112\pm5.8$)} \\
        \midrule 
        \multirow{4.5}{*}{$256\times256$}& \multirow{2}{*}{$\text{ICt}(5)$}  & \multirow{2}{*}{$3.0\cdot10^{-2}\pm7.5\cdot10^{-4}$} & $5.031\pm0.622$ & $6.222\pm0.754$ & $7.178\pm0.852$ & $7.987\pm0.942$ \\
        &  &  & \size{8}{($110\pm4.4$)} & \size{8}{($136\pm4.6$)} & \size{8}{($157\pm4.6$)} & \size{8}{($175\pm4.6$)} \\\cmidrule(r){2-7}
        &  \multirow{2}{*}{PreCor$\big[\text{ICt}(1)\big]$}  & \multirow{2}{*}{$1.4\cdot10^{-2}\pm6.8\cdot10^{-4}$} & $\textbf{2.719}\pm0.381$ & $\textbf{3.378}\pm0.462$ & $\textbf{3.976}\pm0.539$ & $\textbf{4.534}\pm0.608$ \\ 
        &  &  & \size{8}{($105\pm8.3$)} & \size{8}{($131\pm9.8$)} & \size{8}{($154\pm11.2$)} & \size{8}{($176\pm12.4$)} \\
        \bottomrule
    \end{tabular}
\end{table*}

\section{Experiments}
\label{sec:Experiments}
In our approach, we used both IC($0$) and ICt($1$) to train the PreCorrector. In the next section, we will use the following notations:

\begin{itemize}
    \item IC($0$), ICt($1$) and ICt($5$) are classical preconditioners from linear algebra with a corresponding level of fill-in.
    \item PreCor$\big[$IC($0$)$\big]$ and PreCor$\big[$ICt($1$)$\big]$ is PreCorrector with corresponding preconditioners as input.
\end{itemize}

Details of the experimental environment can be found in Appendix~\ref{app:exp_env}.

\paragraph{Metrics}
The main comparison of preconditioners designed with different algorithms are made by comparing total time including preconditioner construction time and the number of CG iterations to achieve a given tolerance. For construction time, we report averaged values over $200$ runs of preconditioner construction and for CG time and iterations we report averaged values over the test set as well as standard deviations for the average values. Construction time for PreCorrector is reported including construction time of classical preconditioners.

The GNNs is chosen to construct preconditioners because it allows preserve the sparsity pattern. Therefore, the algorithmic complexity of using preconditioners (matrix-vector product) is the same when using preconditioners with the same sparsity pattern. This allows a fair evaluation of the quality of neural preconditioners with the same sparsity pattern only in terms of the number of CG iterations. Furthermore, all approaches to IC decomposition with the same sparsity pattern, including the classical ones, compete with each other in terms of construction time, effect on the spectrum (i.e., number of CG iterations) and generalization ability.

\begin{table*}[!h]
    \footnotesize
    \caption{Comparison on Poisson equation: classical algorithm IC($0$), GNN from~\citep{li2023learning} and PreCorrector. Lower is better, the best results are bold. The second best is underlined. Pre-time stands for precomputations time.}
    \label{table:poisson_lietal}
    \centering
    \begin{tabular}{@{}lcccccc@{}} 
        \\\toprule
        &  &  & Time (iters) & Tme (iters) & Time (iters) & Time (iters) \\
        Grid  & Method  & Pre-time &  to $10^{-3}$  &to $10^{-6}$ &to $10^{-9}$ &to $10^{-12}$  \\
        \midrule 
        \multirow{7}{*}{$128\times128$}&  \multirow{2}{*}{$\text{IC}(0)$}  & \multirow{2}{*}{$4.7\cdot10^{-4}\pm2.0\cdot10^{-5}$} & $0.762\pm0.116$ & $1.012\pm0.153$ & $1.154\pm0.175$ & $1.353\pm0.205$ \\
        &  &  & \size{8}{($118\pm0.0$)} & \size{8}{($157\pm0.0$)} & \size{8}{($179\pm0.0$)} & \size{8}{($210\pm0.0$)} \\\cmidrule(r){3-7}
        &  \multirow{2}{*}{PreCor$\big[\text{IC}(0)\big]$}  & \multirow{2}{*}{$8.1\cdot10^{-4}\pm1.2\cdot10^{-4}$} & $\textbf{0.403}\pm0.009$ & $\textbf{0.520}\pm0.009$ & $\textbf{0.620}\pm0.010$ & $\textbf{0.729}\pm0.011$ \\ 
        &  &  & \size{8}{($48\pm0.0$)} & \size{8}{($62\pm0.0$)} & \size{8}{($74\pm0.0$)} & \size{8}{($87\pm0.0$)} \\\cmidrule(r){3-7}
        &  \multirow{2}{*}{\citep{li2023learning}}  & \multirow{2}{*}{$1.1\cdot10^{-3}\pm2.5\cdot10^{-4}$} & $\underline{0.471}\pm0.058$ & $\underline{0.589}\pm0.069$ & $\underline{0.708}\pm0.081$ & $\underline{0.824}\pm0.089$ \\ 
        &  &  & \size{8}{($51\pm0.0$)} & \size{8}{($64\pm0.0$)} & \size{8}{($77\pm0.0$)} & \size{8}{($90\pm0.3$)} \\
        \bottomrule
    \end{tabular}
\end{table*}

\begin{table*}[!h]
    \footnotesize
    \caption{Comparison on diffusion equation with grid $32$ with varying variance in GRF: classical algorithms for IC($0$) and GNN from~\citep{li2023learning}. Lower is better, the best results are bold. Pre-time stands for precomputations time. $1)$None of the test linear systems converged to $10^{-3}$ tolerance in $300$ iterations.}
    \label{table:diff_0.1_lietal}
    \centering
    \begin{tabular}{@{}lcccccc@{}} 
        \\\toprule
        &  &  & Time (iters) & Time (iters) & Time (iters) & Time (iters) \\
        Variance  & Method  & Pre-time &  to $10^{-3}$  &to $10^{-6}$ &to $10^{-9}$ &to $10^{-12}$ \\
        \midrule 
        \multirow{4.5}{*}{$0.1$}&  \multirow{2}{*}{$\text{IC}(0)$}  & \multirow{2}{*}{$7.3\cdot10^{-5}\pm2.7\cdot10^{-6}$} & $\textbf{0.046}\pm0.004$ & $\textbf{0.060}\pm0.005$ & $\textbf{0.071}\pm0.005$ & $\textbf{0.081}\pm0.006$ \\ 
        &  &  & \size{8}{($33\pm0.7$)} & \size{8}{($43\pm0.76$)} & \size{8}{($52\pm0.6$)} & \size{8}{($59\pm0.6$)} \\\cmidrule(r){3-7}
        &  \multirow{2}{*}{\citep{li2023learning}}  & \multirow{2}{*}{$3.1\cdot10^{-4}\pm3.3\cdot10^{-5}$} & $0.081\pm0.008$ & $0.106\pm0.001$ & $0.128\pm0.012$ & $0.150\pm0.014$ \\ 
        &  &  & \size{8}{($61\pm5.9$)} & \size{8}{($80\pm7.4$)} & \size{8}{($97\pm9.2$)} & \size{8}{($114\pm10.8$)} \\
        \midrule
        \multirow{4.5}{*}{$0.5$}&  \multirow{2}{*}{$\text{IC}(0)$}  & \multirow{2}{*}{$7.4\cdot10^{-5}\pm2.5\cdot10^{-6}$} & $\textbf{0.052}\pm0.004$ & $\textbf{0.066}\pm0.005$ & $\textbf{0.078}\pm0.006$ & $\textbf{0.088}\pm0.007$ \\ 
        &  &  & \size{8}{($37\pm1.2$)} & \size{8}{($48\pm1.1$)} & \size{8}{($57\pm1.2$)} & \size{8}{($64\pm1.2$)} \\\cmidrule(r){3-7}
        &  \multirow{2}{*}{\citep{li2023learning}}  & \multirow{2}{*}{$4.1\cdot10^{-4}\pm1.3\cdot10^{-5}$} & \multirow{2}{*}{nan$^{1}$} & \multirow{2}{*}{nan$^{1}$} & \multirow{2}{*}{nan$^{1}$} & \multirow{2}{*}{nan$^{1}$} \\
        &  &  &  &  &  \\
        \bottomrule
    \end{tabular}
\end{table*}

\subsection{Preconditioners comparison}
\label{sec:comparison}

\paragraph{Experiments with classical algorithms}

Preconditioners constructed with PreCorrector outperform classical algorithms with the same sparsity pattern in both total time and effect on the spectrum, i.e. CG iterations (Table~\ref{table:diff_1.1}). The difference between the PreCorrector's inference and the construction of the IC is negligible and, most importantly, contributes a little to the final time-to-solution. At the same time PreCor$\big[$IC($0$)$\big]$ requires fewer iterations to achieve the required tolerance and thus has better effect on the spectrum of $A$ than IC($0$). Since PreCorrector takes $L$ from any classical preconditioner, one can obtain even better preconditioners with more advanced ones. The proposed approach based on the ICt($1$) preconditioner, PreCor$\big[$ICt($1$)$\big]$, also has a better effect on the spectrum of the initial $A$ than the classical ICt($1$).

In Table~\ref{table:diff_1.1} and Table~\ref{table:diff_1.1_ict5} one can observe that PreCor$\big[$IC($0$)$\big]$ and PreCor$\big[$ICt($1$)$\big]$ generally perform on par with ICt($1$) and ICt($5$) respectively in terms of total time, which are denser than the initial left-hand side $A$ (see Appendix~\ref{app:dataset}). In certain experiments these PreCor$\big[$IC($0$)$\big]$ and PreCor$\big[$ICt($1$)$\big]$ have even fewer CG iterations compared to ICt($1$) and ICt($5$) respectively. Consequently, for larger grid sizes, the total time of PreCor$\big[$IC($0$)$\big]$ and PreCor$\big[$ICt($1$)$\big]$ is less than that of ICt($5$). Moreover, the construction cost of the denser preconditioner scales worse than the inference of PreCorrector. Therefore, PreCorrector is a more favourable method for large systems than effective dense preconditioners. Additional information on the scalability of the PreCorrector can be found in the Appendix~\ref{app:scalability}. More results on different datasets can be found in Appendix~\ref{app:ic0_ict1_experiemnts} and Appendix~\ref{app:ict5_experiemnts}.

We observe a good generalization of our approach when transferring our preconditioner between grids and datasets (see Appendix~\ref{app:generalization}). The transferability of the PreCorrector allows both to use the pre-trained PreCorrector on similar problems and to train the PreCorrector on the easier problems with inference on the hard ones.

\begin{figure*}[!h]
\normalsize
\centering
    \begin{center}
        \includegraphics[width=1.\textwidth]{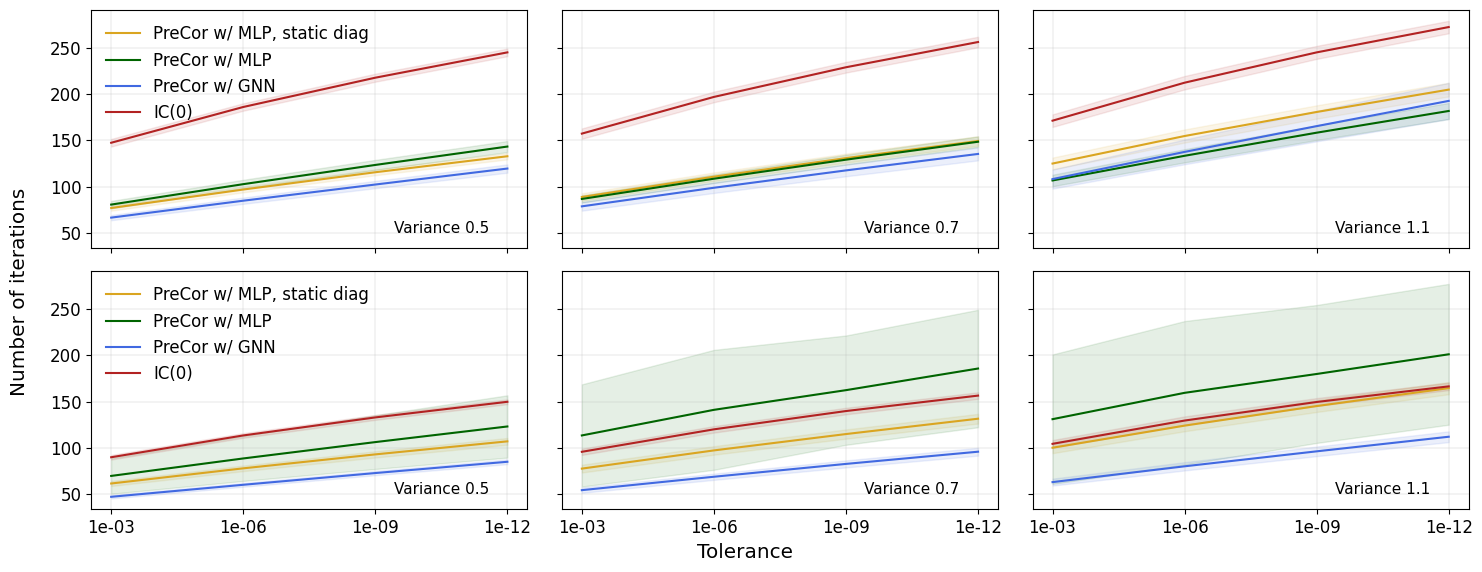}
    \end{center}
    \caption{Ablation study of the PreCorrector's architecture. \textbf{Top row} - preconditioners are constructed from IC($0$). \textbf{Bottom row} - preconditioners are constructed from ICt($1$). PreCor w/ GNN - PreCorrector architecture described in Section~\ref{sec:learn_corrrection_ilu}. PreCor w/ MLP - PreCorrector processor block is changed to MLP. PreCor w/ MLP, static diag - same as before without updating the main diagonal. Lower is better.}
    \label{fig:ablation_precor}
    \vspace*{4pt}
\end{figure*}

\paragraph{Experiments with neural preconditioner design}
Preconditioners constructed with GNNs from previous works~\citep{li2023learning, hausner2023neural} report speedup compared to classical preconditioners when the latter had a long construction time. At the same time, they cannot outperform the effect of IC($0$) on the spectrum of the initial matrix. We implemented the approach of~\citep{li2023learning} and compared it with PreCorrector and IC($0$).

Experiments on simple Poisson equation show that both PreCor$\big[$IC($0$)$\big]$ and GNN from~\citep{li2023learning} have similar number of CG iterations. However, PreCor$\big[$IC($0$)$\big]$ has lighter inference and thus less total time-to-solution. Note that constructing IC($0$) with the \texttt{ilupp} library takes negligible time compared to the total CG time. With PDE coefficients generated as Gaussian random fields with small discontinuities in them, the GNN from~\citep{li2023learning} cannot reach the same number of CG iterations as the classical IC($0$), while quality of the PreCorrector does not degenerate when it is trained on the datasets with higher contrast.

Results for GNN from~\citep{li2023learning} are obtained with the same GNN hyperparameters and architecture as reported in the original paper with only one change: GNN is initialized with all weights and biases equal to zero and trained untill convergence.

Without this change in architecture, the training did not converge. We also observed unstable training of the GNN from~\citep{li2023learning} and unpredictable quality of the preconditioner depending on the training length. Incorporating of a normalization used in PreCorrector did not improve the quality of the GNN. In our experiments, we could get a better preconditioner if we did not train until convergence. However, we could not predict the time for early stopping except by directly computing $\kappa(P^{-1}A)$, which is too expensive.

\subsection{Ablation study}
Learning to correct ILU factors and using a better loss function are key components that allow PreCorrector to produce better preconditioners than classical ones. To further support this claim, we perform an ablation study of these components.

\paragraph{Processor layer of GNN}
We have changed the message-passing processor block of PreCorrector (see Section~\ref{sec:learn_corrrection_ilu}) to a simple MLP, resulting in a feed-forward network without any propagation of neighbourhood information (Figure~\ref{fig:ablation_precor}).

Experiments on the IC($0$) preconditioner show that PreCorrector without message-passing generally produces weaker preconditioners than the architecture described in Section~\ref{sec:learn_corrrection_ilu} and degradates with increasing coefficient contrast. Nevertheless, it achieves the desired generalization on unseen linear systems and outperforms classical IC($0$). On the other hand, the PreCorrector without message-passing cannot handle varying sparsity patterns (experiments with ICt($1$), bottom row, Figure~\ref{fig:ablation_precor}) when such preconditioners are given as input matrix $L$.

By pinning diagonal elements $L_{ii}$ in PreCorrector without message-passing, one can stabilize the learning with varying coefficients at the cost of further performance degradation. Diagonal pinning during training can be seen as a variation of the SPD constraint on the matrix $L(\theta)$ that was first introduced in~\cite{sappl2019deep} and further used in~\cite{li2023learning, hausner2023neural}.

The classical ILU algorithms are non-local by design, which allows PreCorrector to construct better preconditioner than their classical analogues for certain problems even with shallow two-layer MLP with $50$ parameters as the processor supporting the efficiency of the proposed approach. 

\begin{table*}[!h]
    \footnotesize
    \caption{Condition number, spectrum and value of loss~(\ref{eq:loss_minus_A_invA}) for a sampled model from on diffusion equation with variance $0.7$ on grid $128\times128$. The loss value is calculated directly with $A^{-1}$.}
    \centering
    \begin{tabular}{@{}rcrrcrr@{}}
        \\\toprule
        Matrix & $\kappa(P^{-1}A)$ & $ \lambda_{\text{min}}$ & $\lambda_{\text{max}}$ & $\big\Vert LL^\top A^{-1} - I\big\Vert^2_F$ & Bound~$\lambda_{\text{min}}$ & Bound~$\lambda_{\text{max}}$ \\ 
        \midrule
        $A$ & $87565$ & $17.5506$ & $1536813.4$ & --- & --- & --- \\[0.1cm]
        $\big(L_0L_0^{\top}\big)^{-1}A$ & $749$ & $0.0016$ & $1.2$ & $ 1.04\cdot10^{6}$ & $9.7\cdot10^{-4}$ & $230$ \\[0.1cm] 
        $\big(L(\theta)L(\theta)^{\top}\big)^{-1}A$ & $78$ & $0.1981$ & $15.5$ & $ 1.73\cdot10^{3}$ & $8.4\cdot10^{-3}$ & $4157$ \\[0.1cm] 
        \bottomrule
    \end{tabular}
    \label{table:loss_spectrum}
\end{table*}

\begin{table*}[!h]
    \footnotesize
    \caption{Comparison of losses (\ref{eq:loss_minus_A}) and (\ref{eq:loss_minus_A_invA}). Number of CG iterations on the diffusion equation with $0.7$ on grid $64\times64$. During training Hutchinson trick is applied for both losses. $^{1}$Condition number and eigenvalues are calculated on a single sampled linear system.}
    \centering
    \begin{tabular}{@{}crrrccc@{}}
        \\\toprule
        Loss & $10^{-3}$ & $10^{-6}$ & $10^{-9}$ & $\kappa(P^{-1}A)^{1}$ & $\lambda_{\text{min}}^{1}$ & $\lambda_{\text{max}}^{1}$ \\
        \midrule
        (\ref{eq:loss_minus_A}) & $123\pm4.7$ & $155\pm5.3$ & $182\pm5.4$ & $571$ & $0.0033$ & $\textbf{1.88}$ \\
        (\ref{eq:loss_minus_A_invA}) \textbf{(Ours)} & $\textbf{42}\pm1.7$ & $\textbf{55}\pm2.1$ & $\textbf{67}\pm2.5$ & $\textbf{60}$ & $\textbf{0.1406}$ & $8.40$ \\
        \bottomrule
    \end{tabular}
    \label{table:loss_comparison}
\end{table*}

\paragraph{Loss function}
The equivalence of the losses (\ref{eq:loss_minus_A_invA}) and (\ref{eq:loss_with_rhs}) allows to avoid explicit inverse materialization and provides maximum complexity of the matrix-vector product in the loss during training. Recall that $A$ comes from the 5-point finite difference discretization of the diffusion equation (Section~\ref{sec:datasets}). $A$ tends to a diagonal matrix with $n \rightarrow \infty$ and we can assume that $A$ is a diagonal matrix for sufficiently large linear systems. Minimizing a matrix product between the preconditioner and $A^{-1}$ in (\ref{eq:loss_minus_A_invA}) makes the eigenvalues tend to $1$. 

As mentioned in Section~\ref{sec:neural_prec_design}, one should focus on approximating the low frequency components. The loss~(\ref{eq:loss_minus_A_invA}) does indeed reduce the distance between the extreme eigenvalues compared to IC(0) which can be observed in Table~\ref{table:loss_spectrum}. Moreover, the gap between the extreme eigenvalues is covered by the increase in the minimum eigenvalue, which supports the hypothesis of low frequency cancellation. The maximum eigenvalue also increase, but by a much smaller order of magnitude. In Table~\ref{table:loss_spectrum} we have also included recently obtained bounds for the minimum and maximum eigenvalues~\citep{hausner2024learning}.



At the same time, the preconditioner trained with the loss function~(\ref{eq:loss_minus_A}) without weighting with $A^{-1}$ gives a worse effect on spectrum of $A$ (Table~\ref{table:loss_comparison}) and takes much more time to converge (Appendix~\ref{app:losses}). Different distributions of the eigenvalues can be found in Figure~\ref{fig:eigen_distr} in the Appendix~\ref{app:eigen_distr} 

\section{Discussion}

In our work, we propose a neural design of preconditioners for the CG iterative method that can outperform analogous classical preconditioners with the same sparsity pattern of the ILU family in terms of both effect on the spectrum and total time-to-solution. Using the classical preconditioners as a starting point and learning corrections for them, we achieve stable and fast training convergence that can handle parametric PDEs with contrast coefficients. We also propose a complexity metric to measure the complexity of PDEs with random coefficients.

We observe that training a GNN from scratch can be unstable, resulting in preconditioners that have a weaker effect on the spectrum than their classical analogues. In addition, as the matrix size increases, the very first step of training, when GNN is initialized with random weights, leads to an loss overflow since the residual with random $P$ is huge. Learning corrections to classical methods mitigates these problems.

We provide numerical evidence for our observation of low-frequency cancellation with the loss function used. However, we found no trace of this relationship in the numerical analysis literature. We believe that there exists a learnable transformation that will be universal for different sparse matrices to construct the ILU decomposition that will significantly reduce $\kappa(A)$. We propose that this loss analysis is the key ingredient for successful learning of the general form transformation.

\section{Limitations}

The limitations of the proposed approach are as follows:

\begin{itemize}
    \item[1.] Theoretical study of the loss function used. We provide only a heuristic understanding with experimental justification for the loss function. A theoretical analysis of the loss function is the subject of future research.
    \item[2.] The target objective in norms other than $\Vert\cdot \Vert_F$ may provide a tighter bound on the spectrum. Investigating the possible use of target values in other norms is a logical next step.
    \item[3.] Experiments on other meshes and sparsity patterns of the resulting left hand side matrices $A$. Generalization of the PreCorrector to transformation in the space of sparse matrices with general sparsity patterns.
    \item[4.] While the PreCorrector has only been tested on systems with SPD matrices from the discretization of elliptic equations, further work will require generalization to irregular grids, non-symmetric problems, hyperbolic PDEs, nonlinear problems, to other iterative solvers such as GMRES and BiCGSTAB and modification of the preconditioner design accordingly.
    \item[5.] The forcing term $f(x)$ is sampled from the standard normal distribution, but the case of complex forcing terms needs to be studied separately as it can also affect the complexity of solving parametric PDEs.
\end{itemize}

\section*{Impact Statement}

This paper presents work whose goal is to advance the field of 
Machine Learning. There are many potential societal consequences 
of our work, none which we feel must be specifically highlighted here.


\bibliography{example_paper}
\bibliographystyle{icml2025}

\newpage
\appendix
\onecolumn

\section{Appendix}

\subsection{Incomplete LU factorization} 
\label{app:ilu}

Full LU decomposition~\citep{trefethen2022numerical} for a square non-singular matrix defined as the product of the lower and upper triangular matrices $B = L_B U_B$. In general, these matrices have no restriction on the position and number of elements within their triangular structure and can even be dense for a sparse matrix $B$. On the other hand, an ILU is an approximate LU factorization:

\begin{equation}
    \label{eq:ILU0}
    A \approx L U,
\end{equation}

\noindent where $LU - A$ satisfies certain constraints. 

Zero fill-in ILU, denoted ILU($0$), is an approximate LU factorization $A \approx L_0 U_0$ in such a way that $L_0$ has exactly the same sparsity pattern as the lower part of $A$ and $U_0$ has exactly the same sparsity pattern as the upper part of $A$. For the ILU($p$) decomposition the level of fill-in is defined hierarchically. The product of the factors of the ILU($0$) decomposition produces a new matrix $B$ with a larger number of non-zero elements. The factors of the ILU(1) factorization have the same sparsity patterns as lower and upper parts of the sparsity pattern of the matrix $B$. With this recursion one gets a pattern of the ILU($p$) factorization with $p$-level of fill-in. ILU($0$) is a typical choice to precondition iterative solvers and relies only on the levels of fill-in, e.g. sparsity patterns~\citep{saad2003iterative}. One can obtain better approximation with ILU by using incomplete factorizations with thresholding.

One such technique is the ILU factorization with thresholding (ILUt($p$)). The parameter $p$ defines the number of additional non-zeros allowed per column in the resulting factorization. For the ILUt($p$) decomposition, the algorithm is more complex and involves both dropping values by some predefined threshold and controlling the number of possible non-zero values in the factorization. In the case of ILUt($p$), the value $p$ represents additional non-zero values allowed in the factorization per row. The thresholding algorithm provides a more flexible and effective way to approximate the inverse of a matrix, especially for realistic problems where the numerical values of the matrix elements are important.

The complexity of solving sparse linear systems with matrices in the form of the Cholesky decomposition defined by the number of non-zero elements $\mathcal{O}(\text{nnz})$. This value also defines the storage complexity and the complexity of preconditioner construction.

\subsection{PreCorrector's architecture}
\label{app:architecture}

\begin{algorithm}[!h]
   \caption{PreCorrector's forward pass}
   \label{alg:precorrector_forward}
\begin{algorithmic}
   \STATE {\bfseries Input:} IC($0$) factor $L$ of $A$, where $e_{ij}^0 = L_{ij}$
   \STATE $\text{s} = \max{(|e_{ij}^0|)}$
   \STATE $e_{ij}^1 = \gamma_{\theta}(e_{ij}^0~/~\text{s})$
   \STATE $h_i^1 = \textbf{1}$
   \FOR{$t=1$ {\bfseries to} $T$}
   \STATE $h_i^{t+1}=\bigoplus_{j \in \mathcal{N}(i)}h_{j}^t~e_{ij}^t$
   \STATE $e_{ij}^{t+1}=\phi_{\theta}\big(e_{ij}^t, h_i^{t+1}, h_j^{t+1}\big)$
   \ENDFOR
   \STATE $e_{ij}^{T+1} = \text{s}\cdot\psi_{\theta}(e_{ij}^{T+1}) $
   \STATE $e_{ij}^{T+1} = e_{ij}^0 + \alpha\cdot e_{ij}^{T+1}$
   \STATE {\bfseries Return:} $L(\theta)$, where $L(\theta)_{ij} = e_{ij}^{T+1}$
\end{algorithmic}
\end{algorithm}

By using \textbf{1} as an input nodes to the processor block, we discard the node encoder MLP. We also notice that node model in processor does not contribute when there are no explicit nodes. We discard it either to use simple aggregation as the node update model and to further reduce number of parameters. Moreover, we make our approach consistent with classical methods by passing only the left-side matrix $A$ to the preconditioner construction routine. We use the \texttt{max} aggregation function, although our experiments showed that the resulting preconditioner quality does not depend on the type of aggregation function.

We also investigate which protocol to follow to increase the receptive field. It is possible to follow the original message passing paper~\cite{gilmer2017neural} and perform $T$ rounds of message passing with shared learnable functions during each round. Another option is to follow classic GNN's multi-block protocol and combine multiple message-passing layers to increase the receptive field with independent weights. During the PreCorrecotr ablation study we observed no difference in resulting preconditioner between the protocols, so we use message-passing with shared layer to further simplify the model.

\begin{table}[!h]
    \footnotesize
    \caption{Number of learning parameters for different neural preconditioner routines. PreCor - PreCorrector's architecture presented in Algorithm~\ref{alg:precorrector_forward}. PreCor w/ nodes - PreCorrector with node encoder and with node MLP model in Processor. PreCor MB - PreCorrector with multi-block message-passing protocol. PreCor w/o MP - PreCorrector without message-passing (with single MLP).}
    \label{table:models_parameters}
    \centering
    \begin{tabular}{@{}cccc@{}} 
        \\\toprule
        PreCor & PreCor w/ nodes & PreCor MB & PreCor w/o MP \\\cmidrule(r){1-4}
        $1170$ & $2753$ & $3474$ & $50$ \\
        \bottomrule
    \end{tabular}
\end{table}

\subsection{Dataset description}
\label{app:dataset}

For our experiments we use a following list of computational grids: $\{32,~64,~128,~256\}$. The contrast in the diffusion equation is controlled with a variance in the coefficient function GRF and takes value from $\{0.1,~0.5,~0.7,~1.1\}$. The Gaussian covariance model is used in GRF. The forcing term $f$ is sampled from the standard normal distribution $\mathcal{N}(0,~1)$ and each PDE is discretized using the 5-point finite difference method.

\begin{figure*}[!h]
\normalsize
\centering
    \begin{center}
        \includegraphics[width=1.0\textwidth]{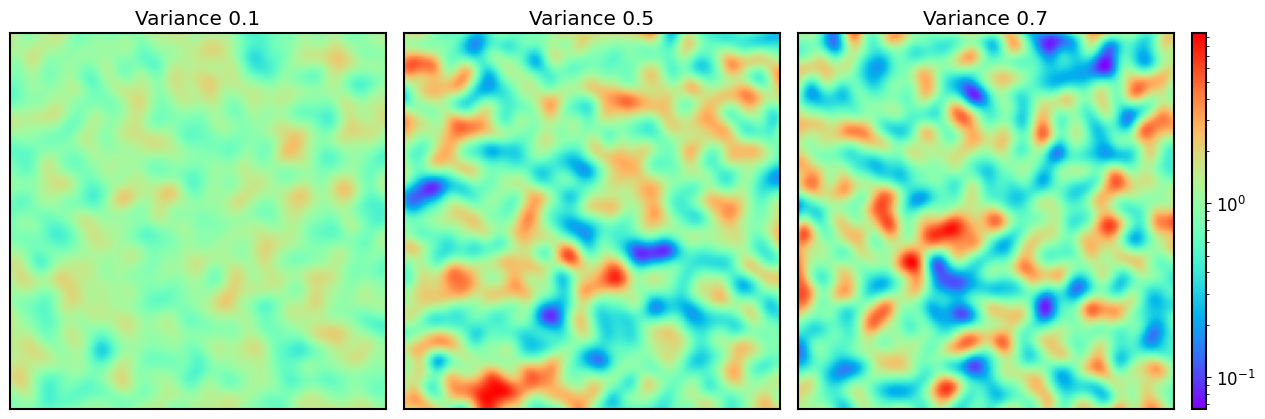}
    \end{center}
    \caption{Coefficient function $k(x) = \exp{(\phi(x))}$ for grid $128\times128$ with different variances.}
    \label{fig:coeff}
    \vspace*{4pt}
\end{figure*}

\begin{table}[!h]
    \caption{Contrast values for diffusion equation with various variances.}
    \centering
    \begin{tabular}{@{}lcccc@{}}
        \\\toprule 
        Variance & Grid  & $\text{Min}~\text{contrast}$ & $\text{Mean} ~\text{contrast}$ & $\text{Max}~\text{contrast}$ \\
        \midrule
        \multirow{3}{*}{$0.1$} & $32$ & $5$ & $7$ & $11$ \\
        & $64$ & $5$ & $8$ & $12$ \\
        & $128$ & $6$ & $8$ & $14$ \\
        \midrule
        \multirow{4}{*}{$0.5$} & $32$ & $36$ & $86$ & $179$ \\
        & $64$ & $45$ & $103$ & $200$ \\
        & $128$ & $50$ & $116$ & $297$ \\
        & $256$ & $81$ & $127$ & $229$ \\
        \midrule
        \multirow{4}{*}{$0.7$}& $32$ & $180$ & $277$ & $697$ \\
        & $64$ & $200$ & $318$ & $742$ \\
        & $128$ & $300$ & $426$ & $798$ \\
        & $256$ & $251$ & $389$ & $879$ \\
        \midrule
        \multirow{3}{*}{$1.1$}& $64$ & $404$ & $768$ & $1199$ \\
        & $128$ & $400$ & $790$ & $1190$ \\
        & $256$ & $404$ & $803$ & $1200$ \\
        \bottomrule 
    \end{tabular}
    \label{table:dataset_specification}
\end{table}

\begin{table}[!h]
    \caption{Size of linear systems and number of nonzero elements (nnz) for different grid sizes and matrices.}
    \centering
    \begin{tabular}{@{}ccccccccc@{}}
        \\\toprule
        & \multicolumn{2}{c}{Grid $32\times32$} & \multicolumn{2}{c}{Grid $64\times64$} & \multicolumn{2}{c}{Grid $128\times128$} & \multicolumn{2}{c}{Grid $256\times256$} \\\cmidrule(r){2-3} \cmidrule(r){4-5} \cmidrule(r){6-7} \cmidrule(r){8-9}
        Matrix & Size & nnz, $\%$ & Size & nnz, $\%$ & Size & nnz, $\%$ & Size & nnz, $\%$ \\
        \midrule
        $A$ & \multirow{4}{*}{$1024$} & 0.4761 & \multirow{4}{*}{$4096$} & $0.1205$ & \multirow{4}{*}{$16384$} & $0.0303$ & \multirow{4}{*}{$65536$} & $0.0076$ \\  
        $L$ from IC($0$) &  & $0.2869$ &  & $0.0725$ &  & $0.0182$ &  & $0.0046$ \\  
        $L$ from ICt($1$) &  & $0.3785$ &  & $0.0961$ &  & $0.0242$ &  & $0.0061$ \\  
        $L$ from ICt($5$) &  & $0.7547$ &  & $0.1920$ &  & $0.0485$ &  & $0.0121$ \\  
        \bottomrule
    \end{tabular}
    \label{table:precs_size}
\end{table}

\begin{table}[!h]
    \caption{Condition number of a sampled linear systems from the used datasets.}
    \centering
    \begin{tabular}{@{}ccccc@{}}
        \toprule 
        & & \multicolumn{3}{c}{Grid} \\
        & & $64$ & $128$ & $256$ \\
        \midrule
        \multirow{3.5}{*}{Variance} & $0.5$ & $9.9 \cdot 10^{3}$ & $5.8 \cdot 10^{4}$ & $ 3.5 \cdot 10^{5}$ \\
        \cmidrule(r){3-5}
        & $0.7$ & $2.7 \cdot 10^{4}$ & $8.8 \cdot 10^{4}$ & $6.0 \cdot 10^{5}$ \\
        \cmidrule(r){3-5}
        & $1.1$ & $42 \cdot 10^{4}$ & $1.8 \cdot 10^{5}$ & $6.4 \cdot 10^{5}$ \\
        \bottomrule 
    \end{tabular}
    \label{table:cond}
\end{table}

\subsection{Experiments environment}
\label{app:exp_env}
Each dataset from the Section~\ref{sec:datasets} consists of $1000$ training and $200$ test linear systems. The final neural networks are trained with batch size $8$, learning rate $10^{-3}$ and Adam optimizer. Final GNN architecture consists of $5$ message passing rounds and $2$ hidden layers with $16$ hidden features in all MLPs in each experiment. PreCorrector training always starts with the parameter $\alpha = 0$ in~(\ref{eq:correct_ic0}). For training, we used libraries from the JAX ecosystem: \texttt{jax}~\citep{jax2018github}, \texttt{optax}~\citep{deepmind2020jax}, \texttt{equinox}~\citep{kidger2021equinox}. We used a single GPU Nvidia A40 48Gb for training. The construction time of preconditioners with neural design was measured on the same GPU. Preconditioners with classical algorithms were generated on the Intel(R) Xeon(R) Gold 6342 CPU @ 2.80GHz with \texttt{ilupp} library\footnote{https://github.com/c-f-h/ilupp}~\cite{mayer2007multilevel}. The CG method was run on the same CPU using the \texttt{scipy}~\citep{2020SciPy-NMeth} implementation.

\subsection{Scalability}
\label{app:scalability}

The PreCorrector is a combination of several MLPs with message-passing connections. When growing non-zero elements nnz (i.e., growing a matrix size) with the same sparsity pattern, the complexity of the PreCorrector forward call grows linearly. The algorithm for IC(0) has $\mathcal{O}(\text{nnz})$. We expect a small increase in computational cost for larger systems, since with matrix grows we are only interested in the growth of non-zero elements in the matrix. The growth of non-zero elements is much less severe than the growth of a matrix size due to the nature of matrices – discretization of PDEs. The PreCorrector works directly on the edges and nodes of the graph and can be easily data-parallelized.

\break 

\subsection{Additional experiments with IC(0) and ICt(1) preconditioners}
\label{app:ic0_ict1_experiemnts}

\begin{figure*}[!h]
\normalsize
\centering
    \begin{center}
        \includegraphics[width=1.\textwidth]{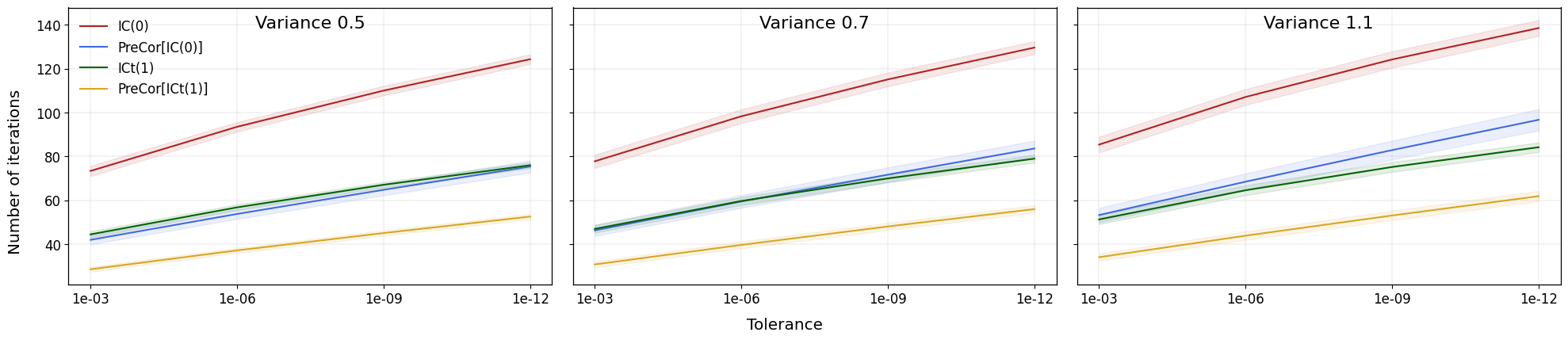}
    \end{center}
    \caption{Residuals vs CG iterations for grid $64\times64$. Lower is better.}
    \label{fig:dkg_64_iters}
    \vspace*{4pt}
\end{figure*}

\begin{figure*}[!h]
\normalsize
\centering
    \begin{center}
        \includegraphics[width=1.\textwidth]{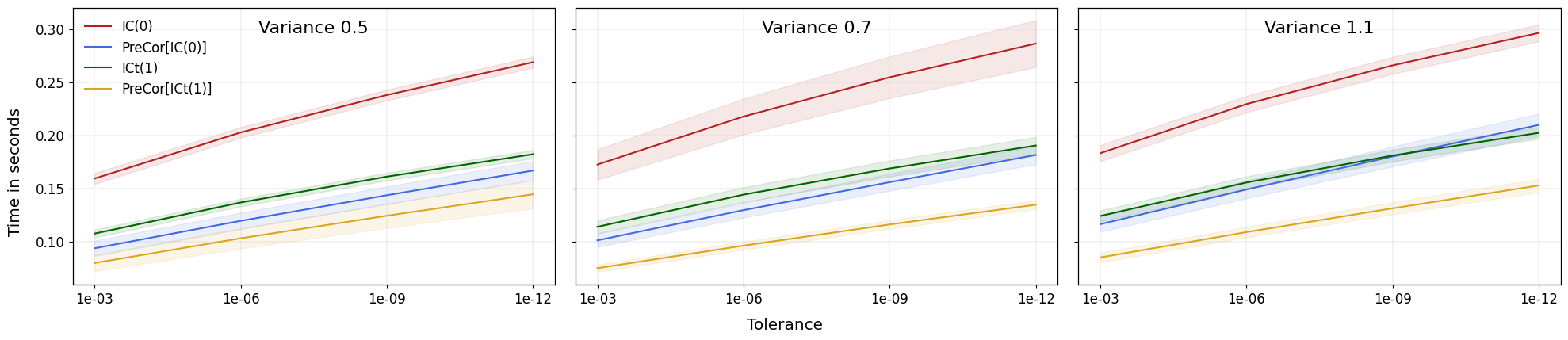}
    \end{center}
    \caption{Wall time vs CG iterations for grid $64\times64$. Lower is better.}
    \label{fig:dkg_64_time}
    \vspace*{4pt}
\end{figure*}

\begin{figure*}[!h]
\normalsize
\centering
    \begin{center}
        \includegraphics[width=1.\textwidth]{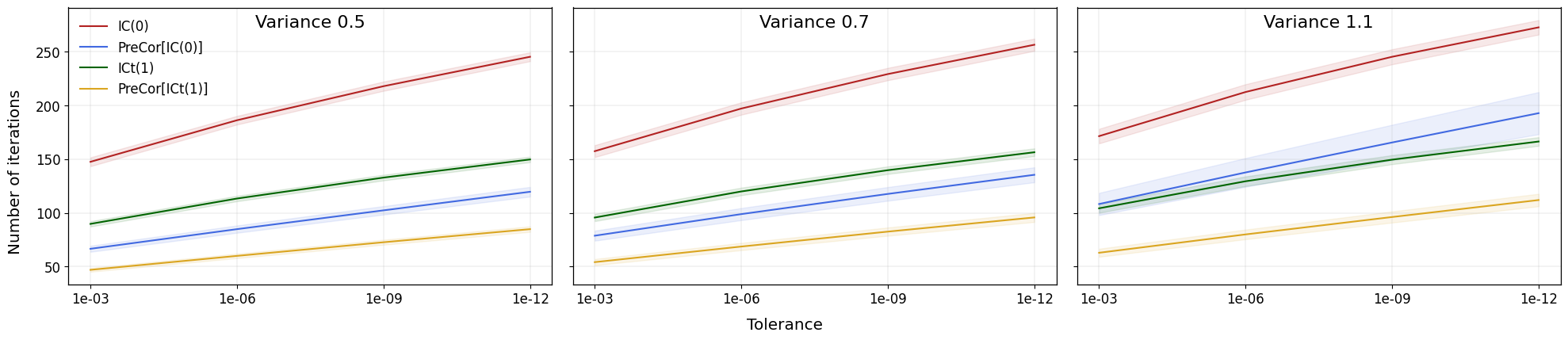}
    \end{center}
    \caption{Residuals vs CG iterations for grid $128\times128$. Lower is better.}
    \label{fig:dkg_128_iters}
    \vspace*{4pt}
\end{figure*}

\begin{figure*}[!h]
\normalsize
\centering
    \begin{center}
        \includegraphics[width=1.\textwidth]{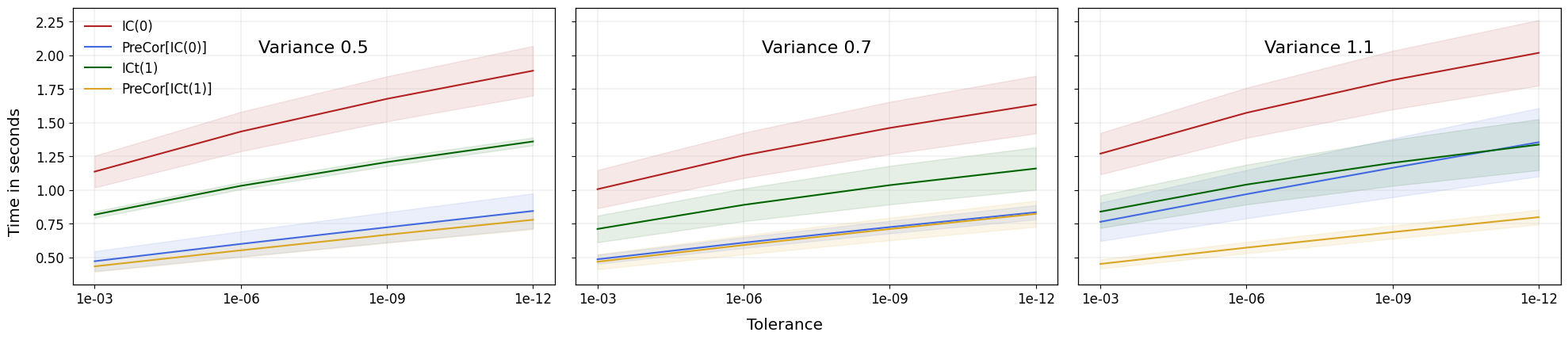}
    \end{center}
    \caption{Wall time vs CG iterations for grid $128\times128$. Lower is better.}
    \label{fig:dkg_128_time}
    \vspace*{4pt}
\end{figure*}

\begin{figure*}[!h]
\normalsize
\centering
    \begin{center}
        \includegraphics[width=1.\textwidth]{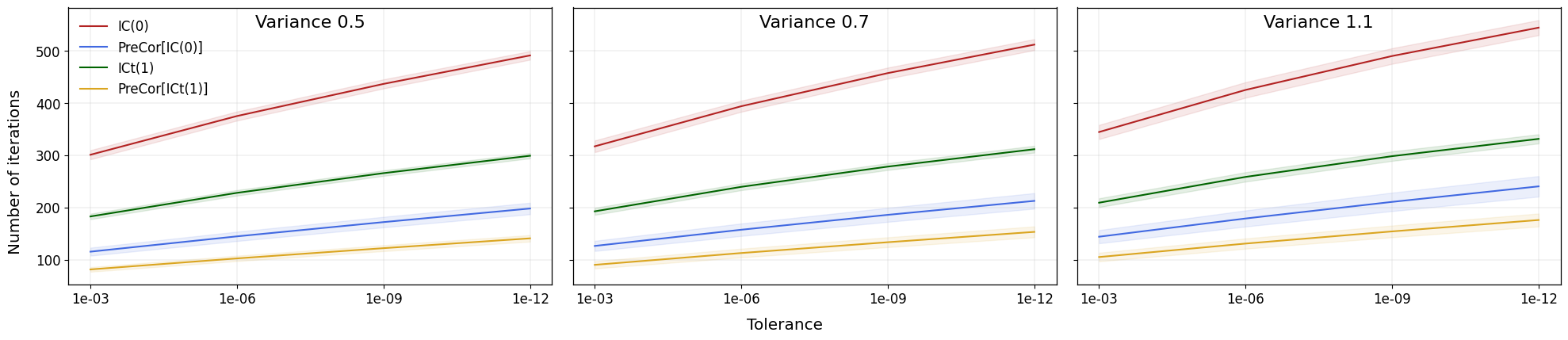}
    \end{center}
    \caption{Residuals vs CG iterations for grid $256\times256$. Lower is better.}
    \label{fig:dkg_256_iters}
    \vspace*{4pt}
\end{figure*}

\begin{figure*}[!h]
\normalsize
\centering
    \begin{center}
        \includegraphics[width=1.\textwidth]{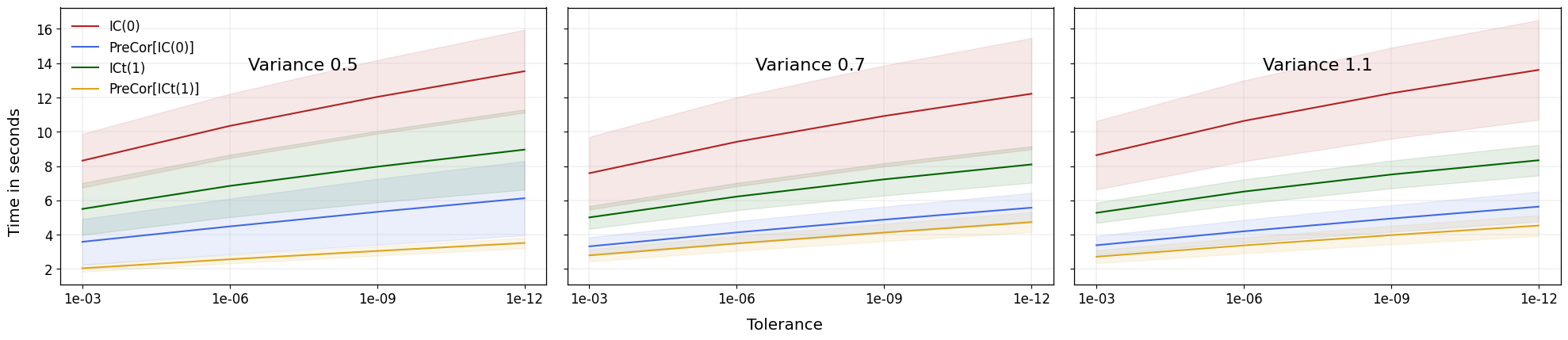}
    \end{center}
    \caption{Wall time vs CG iterations for grid $256\times256$. Lower is better.}
    \label{fig:dkg_256_time}
    \vspace*{4pt}
\end{figure*}

\break 

\subsection{Additional experiments with ICt(5) preconditioner}
\label{app:ict5_experiemnts}

\begin{table}[!h]
    \footnotesize
    \caption{Comparison on diffusion equation with variance $0.5$: classical algorithm ICt($5$) and PreCor$\big[\text{ICt}(1)\big]$. Pre-time stands for precomputations time.}
    \label{table:diff_0.1_ict5_var05}
    \centering
    \begin{tabular}{@{}lcccccc@{}} 
        \\\toprule
        &  &  & Time (iters) & Time (iters) & Time (iters) & Time (iters)  \\
        Grid  & Method  & Pre-time &  to $10^{-3}$  &to $10^{-6}$ &to $10^{-9}$ &to $10^{-12}$  \\
        \midrule
        \multirow{4.5}{*}{$64\times64$}& \multirow{2}{*}{$\text{ICt}(5)$}  & \multirow{2}{*}{$2.1\cdot10^{-3}\pm2.2\cdot10^{-4}$} & $\textbf{0.078}\pm0.003$ & $\textbf{0.099}\pm0.003$ & $\textbf{0.118}\pm0.003$ & $\textbf{0.134}\pm0.003$ \\ 
        &  &  & \size{8}{($22\pm0.7$)} & \size{8}{($28\pm0.6$)} & \size{8}{($34\pm0.6$)} & \size{8}{($39\pm0.7$)} \\\cmidrule(r){2-7}
        &  \multirow{2}{*}{PreCor$\big[\text{ICt}(1)\big]$}  & \multirow{2}{*}{$1.0\cdot10^{-3}\pm2.6\cdot10^{-5}$} & $0.080\pm0.008$ & $0.103\pm0.010$ & $0.124\pm0.012$ & $0.145\pm0.013$ \\ 
        &  &  & \size{8}{($29\pm1.1$)} & \size{8}{($37\pm1.2$)} & \size{8}{($45\pm1.2$)} & \size{8}{($53\pm1.2$)} \\
        \midrule 
        \multirow{4.5}{*}{$128\times128$}& \multirow{2}{*}{$\text{ICt}(5)$}  & \multirow{2}{*}{$7.9\cdot10^{-3}\pm5.6\cdot10^{-4}$} & $0.582\pm0.066$ & $0.737\pm0.082$ & $0.866\pm0.096$ & $0.977\pm0.108$ \\
        &  &  & \size{8}{($46\pm1.4$)} & \size{8}{($59\pm1.4$)} & \size{8}{($69\pm1.3$)} & \size{8}{($78\pm1.3$)} \\\cmidrule(r){2-7}
        &  \multirow{2}{*}{PreCor$\big[\text{ICt}(1)\big]$}  & \multirow{2}{*}{$3.6\cdot10^{-3}\pm1.9\cdot10^{-4}$} & $\textbf{0.432}\pm0.040$ & $\textbf{0.552}\pm0.051$ & $\textbf{0.700}\pm0.060$ & $\textbf{0.788}\pm0.070$ \\ 
        &  &  & \size{8}{($47\pm1.8$)} & \size{8}{($60\pm2.1$)} & \size{8}{($73\pm2.3$)} & \size{8}{($85\pm2.6$)} \\
        \midrule
        \multirow{4.5}{*}{$256\times256$}& \multirow{2}{*}{$\text{ICt}(5)$}  & \multirow{2}{*}{$3.0\cdot10^{-2}\pm1.6\cdot10^{-3}$} & $4.776\pm0.410$ & $5.958\pm0.507$ & $6.942\pm0.583$ & $7.817\pm0.650$ \\
        &  &  & \size{8}{($97\pm2.7$)} & \size{8}{($121\pm2.9$)} & \size{8}{($142\pm2.8$)} & \size{8}{($160\pm2.6$)} \\\cmidrule(r){2-7}
        &  \multirow{2}{*}{PreCor$\big[\text{ICt}(1)\big]$}  & \multirow{2}{*}{$1.4\cdot10^{-2}\pm6.9\cdot10^{-4}$} & $\textbf{2.043}\pm0.203$ & $\textbf{2.567}\pm0.241$ & $\textbf{3.053}\pm0.273$ & $\textbf{3.519}\pm0.303$ \\ 
        &  &  & \size{8}{($82\pm4.5$)} & \size{8}{($103\pm5.2$)} & \size{8}{($122\pm5.7$)} & \size{8}{($141\pm6.2$)} \\
        \bottomrule
    \end{tabular}
\end{table}

\begin{table}[!h]
    \footnotesize
    \caption{Comparison on diffusion equation with variance $0.7$: classical algorithm ICt($5$) and PreCor$\big[\text{ICt}(1)\big]$. Pre-time stands for precomputations time.}
    \label{table:diff_0.1_ict5_var07}
    \centering
    \begin{tabular}{@{}lcccccc@{}} 
        \\\toprule
        &  &  & Time (iters) & Time (iters) & Time (iters) & Time (iters)  \\
        Grid  & Method  & Pre-time &  to $10^{-3}$  &to $10^{-6}$ &to $10^{-9}$ &to $10^{-12}$  \\
        \midrule
        \multirow{4.5}{*}{$64\times64$}& \multirow{2}{*}{$\text{ICt}(5)$}  & \multirow{2}{*}{$2.1\cdot10^{-3}\pm1.9\cdot10^{-4}$} & $\textbf{0.080}\pm0.003$ & $\textbf{0.102}\pm0.003$ & $\textbf{0.120}\pm0.003$ & $\textbf{0.136}\pm0.003$ \\ 
        &  &  & \size{8}{($23\pm0.8$)} & \size{8}{($29\pm0.9$)} & \size{8}{($34\pm0.8$)} & \size{8}{($39\pm0.8$)} \\\cmidrule(r){2-7}
        &  \multirow{2}{*}{PreCor$\big[\text{ICt}(1)\big]$}  & \multirow{2}{*}{$1.4\cdot10^{-3}\pm2.7\cdot10^{-5}$} & $0.075\pm0.003$ & $0.096\pm0.004$ & $0.116\pm0.004$ & $0.135\pm0.004$ \\ 
        &  &  & \size{8}{($31\pm1.4$)} & \size{8}{($40\pm1.6$)} & \size{8}{($48\pm1.7$)} & \size{8}{($56\pm1.7$)} \\
        \midrule 
        \multirow{4.5}{*}{$128\times128$}& \multirow{2}{*}{$\text{ICt}(5)$}  & \multirow{2}{*}{$7.5\cdot10^{-3}\pm4.2\cdot10^{-4}$} & $0.661\pm0.053$ & $0.829\pm0.065$ & $0.968\pm0.075$ & $1.088\pm0.084$ \\
        &  &  & \size{8}{($49\pm1.7$)} & \size{8}{($61\pm1.7$)} & \size{8}{($71\pm1.8$)} & \size{8}{($80\pm1.7$)} \\\cmidrule(r){2-7}
        &  \multirow{2}{*}{PreCor$\big[\text{ICt}(1)\big]$}  & \multirow{2}{*}{$3.6\cdot10^{-3}\pm1.9\cdot10^{-4}$} & $\textbf{0.470}\pm0.056$ & $\textbf{0.591}\pm0.070$ & $\textbf{0.710}\pm0.084$ & $\textbf{0.823}\pm0.097$ \\ 
        &  &  & \size{8}{($54\pm2.9$)} & \size{8}{($69\pm3.4$)} & \size{8}{($83\pm3.8$)} & \size{8}{($96\pm4.2$)} \\
        \midrule
        \multirow{4.5}{*}{$256\times256$}& \multirow{2}{*}{$\text{ICt}(5)$}  & \multirow{2}{*}{$3.0\cdot10^{-2}\pm1.6\cdot10^{-3}$} & $4.800\pm0.624$ & $5.980\pm0.764$ & $6.938\pm0.874$ & $7.776\pm0.968$ \\
        &  &  & \size{8}{($102\pm3.4$)} & \size{8}{($127\pm3.4$)} & \size{8}{($147\pm3.3$)} & \size{8}{($165\pm3.3$)} \\\cmidrule(r){2-7}
        &  \multirow{2}{*}{PreCor$\big[\text{ICt}(1)\big]$}  & \multirow{2}{*}{$1.4\cdot10^{-2}\pm6.8\cdot10^{-4}$} & $\textbf{2.800}\pm0.357$ & $\textbf{3.491}\pm0.434$ & $\textbf{4.126}\pm0.509$ & $\textbf{4.735}\pm0.576$ \\ 
        &  &  & \size{8}{($90\pm7.0$)} & \size{8}{($113\pm8.3$)} & \size{8}{($134\pm9.4$)} & \size{8}{($154\pm10.4$)} \\
        \bottomrule
    \end{tabular}
\end{table}

\break

\subsection{Losses comparison}
\label{app:losses}

\begin{figure*}[!h]
\normalsize
\centering
    \begin{center}
        \includegraphics[width=1.\textwidth]{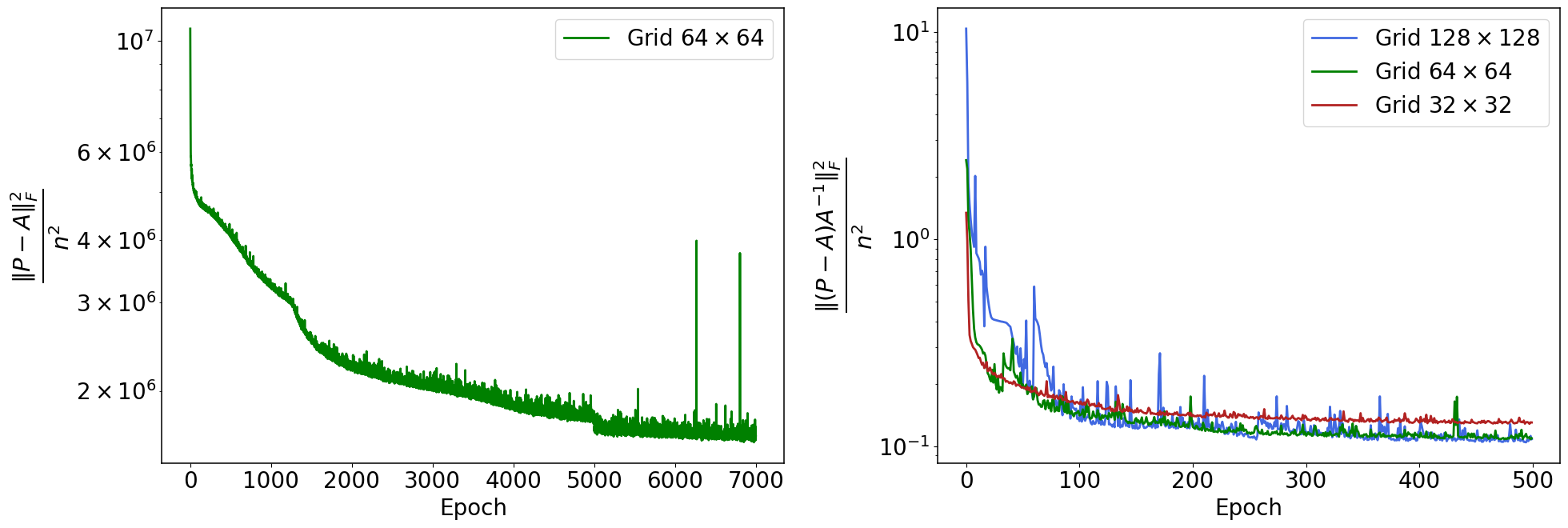}
    \end{center}
    \caption{Test losses during training of PreCor$\big[$IC(0)$\big]$ on the diffusion equation with variance $0.7$ During training Hutchinson trick is applied for both losses.}
    \label{fig:loss}
    \vspace*{4pt}
\end{figure*}

\break

\subsection{Distribution of eigenvalues}
\label{app:eigen_distr}

\begin{figure*}[!h]
\normalsize
\centering
    \begin{center}
        \includegraphics[width=1.0\textwidth]{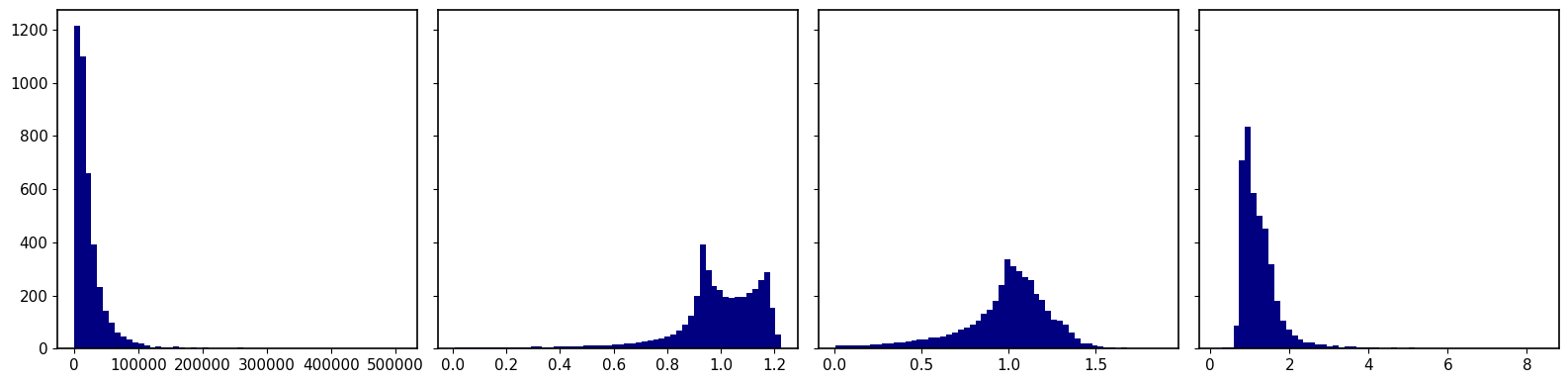}
    \end{center}
    \caption{Distribution of eigenvalues for a sampled linear system of diffusion equation with variance $0.7$ on grid $64\times64$. (\textbf{From left to right}): initial left hand side $A$, $A$ preconditioned with IC($0$), $A$ preconditioned with PreCor$\big[\text{IC}(0)\big]$ trained with loss~(\ref{eq:loss_minus_A}), $A$ preconditioned with PreCor$\big[\text{IC}(0)\big]$ trained with loss~(\ref{eq:loss_minus_A_invA}). Hutchinson estimator is used for both losses.}
    \label{fig:eigen_distr}
    \vspace*{4pt}
\end{figure*}

\break

\subsection{Generalization}
\label{app:generalization}

\begin{figure*}[!h]
\normalsize
\centering
    \begin{center}
        \includegraphics[width=0.9\textwidth]{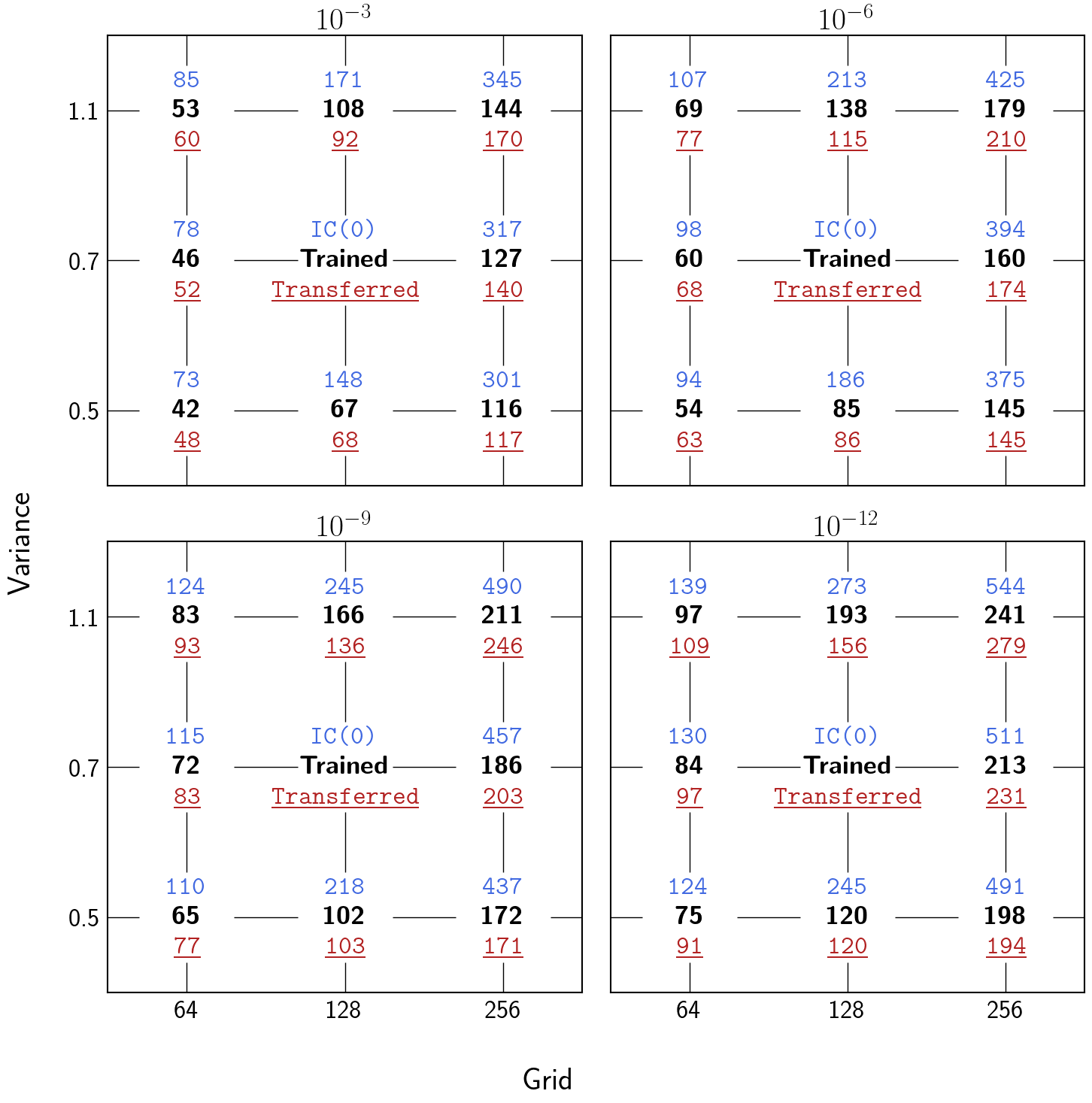}
    \end{center}
    \caption{Generalization of PreCorrector on unseen datasets. Values are number of CG iterations to achieve required tolerance, which specified in the title of each plot. \textbf{Blue} -- IC(0) is used as preconditioner. \textbf{Black} -- PreCorrector$\big[$IC(0)$\big]$; trained and inferenced on the same dataset. \textbf{Red} -- PreCorrector$\big[$IC(0)$\big]$; trained on the diffusion eqaution with variance $0.5$ on grid $64\times64$; inferenced on the dataset, that is described by axes values.}
    \label{fig:generalization}
    \vspace*{4pt}
\end{figure*}




\end{document}